\documentclass[11pt,a4paper]{article}
\usepackage[hyperref]{emnlp2020}
\usepackage{latexsym}
\usepackage{lmodern}
\usepackage{latexsym}
\usepackage{amsmath}
\usepackage{amssymb}
\usepackage{cleveref}
\usepackage{graphicx}
\usepackage{booktabs}
\usepackage{parsetree}
\usepackage[ruled,vlined,noresetcount]{algorithm2e}
\usepackage{tikz}
\usepackage{tikz-qtree}
\usepackage[linguistics]{forest}
\usepackage{subcaption}
\usepackage{times}

\usepackage{microtype}

\crefformat{section}{\S#2#1#3} 
\crefformat{subsection}{\S#2#1#3}
\crefformat{subsubsection}{\S#2#1#3}

\definecolor{dkgreen}{RGB}{0,130,0}

\aclfinalcopy 

\newcommand{\interalia}[1]{\citep[\emph{inter alia}]{#1}}

\newcommand{\na}{\textsc{n/a}}
\newcommand{\st}{\textsc{st}}
\newcommand{\sub}{\textsc{sub}}

\title{On the Role of Supervision in Unsupervised Constituency Parsing}

\author{Haoyue Shi \qquad Karen Livescu \qquad Kevin Gimpel \\
  Toyota Technological Institute at Chicago, IL, USA, 60637 \\
  \texttt{\{freda,klivescu,kgimpel\}@ttic.edu} 
 }

\date{}

\begin{document}
\maketitle
\begin{abstract}
We analyze several recent unsupervised constituency parsing models, which are tuned with respect to the parsing $F_1$ score on the \textit{Wall Street Journal} (WSJ) development set (1,700 sentences).
We introduce strong baselines for them, by training an existing supervised 
parsing model \citep{kitaev-klein-2018-constituency} on the same labeled examples they access. 
When training on the 1,700 examples, or even when using only 50 examples for training and 5 for development, such a few-shot parsing approach can outperform all the unsupervised parsing methods by a significant margin. 
Few-shot parsing can be further improved by a simple data augmentation method and self-training. 
This suggests that, in order to arrive at fair conclusions, we should carefully consider the amount of labeled data used for model development.
We propose two protocols for future work on unsupervised parsing: (i) use fully unsupervised criteria for hyperparameter tuning and model selection; (ii) use as few labeled examples as possible for model development, and compare to few-shot parsing trained on the same labeled examples.\footnote{~Project page: \url{https://ttic.uchicago.edu/~freda/project/rsucp/}}
\end{abstract}

\section{Introduction}
Recent work has considered neural unsupervised constituency parsing \interalia{shen2018neural,drozdov-etal-2019-unsupervised,kim-etal-2019-unsupervised}, showing that it can achieve much better performance than trivial baselines. 
However, many of these approaches use the gold parse trees of all sentences in a development set for either early stopping \interalia{shen2018neural,shen2019ordered,drozdov-etal-2019-unsupervised} or hyperparameter tuning \citep{kim-etal-2019-compound}. 
In contrast, models trained and tuned without any labeled data \citep{kim-etal-2019-unsupervised,peng-etal-2019-palm} are much less competitive. 

Are the labeled examples important in order to obtain decent unsupervised parsing performance?
How well can we do if we \emph{train} on these labeled examples rather than merely using them for tuning?
In this work, we consider training a supervised constituency parsing model  \citep{kitaev-klein-2018-constituency} with very few examples as a strong baseline for unsupervised parsing tuned on labeled examples. 

We empirically characterize unsupervised and few-shot parsing across the spectrum of labeled data availability, finding that (i) tuning based on a few (as few as 15) labeled examples is sufficient to improve unsupervised parsers over fully unsupervised criteria by a significant margin; (ii) unsupervised parsing with supervised tuning does outperform few-shot parsing with fewer than 15 labeled examples, but few-shot parsing quickly dominates once there are more than 55 examples; and (iii) when few-shot parsing is combined with a simple data augmentation method and self-training \interalia{steedman-etal-2003-bootstrapping,reichart-rappoport-2007-self,mcclosky-etal-2006-effective}, only 15 examples are needed for few-shot parsing to begin to dominate. 

Based on these results, we propose the following two protocols for future work on unsupervised parsing: \\[-0.6cm]
\begin{enumerate}
\item Derive and use fully unsupervised criteria for hyperparameter tuning and model selection. \\[-0.6cm]
\item Use as few labeled examples as possible for model development and tuning, and compare to few-shot parsing models trained on the used examples as a strong baseline. \\[-0.6cm]
\end{enumerate}
We suggest future work to tune and compare models under each protocol separately.

In addition, we present two side findings on unsupervised parsing: (i) the vocabulary size in unsupervised parsing, which has not been widely considered as a hyperparameter and varies across prior work, greatly affects the performance of all unsupervised parsing models tested; and (ii) self-training can help improve all investigated unsupervised parsing \citep{shen2018neural,shen2019ordered,drozdov-etal-2019-unsupervised,kim-etal-2019-compound} and few-shot parsing models, and thus can be considered as a post-processing step in future work. 
 
\section{Related Work}
\paragraph{Unsupervised parsing.} 
During the past two decades, there has been a lot of work on both unsupervised constituency parsing \interalia{klein-manning-2002-generative,klein-manning-2004-corpus,bod-2006-subtrees,bod-2006-unsupervised,seginer-2007-fast,snyder-etal-2009-unsupervised} and unsupervised dependency parsing \interalia{klein-manning-2004-corpus,smith-eisner-2006-annealing,spitkovsky-etal-2011-unsupervised,spitkovsky-etal-2013-breaking}. 
Recent work has proposed several effective models for unsupervised or distantly supervised constituency parsing, optimizing either a language modeling objective \interalia{shen2018neural,shen2019ordered,kim-etal-2019-unsupervised,kim-etal-2019-compound} or other downstream semantic objectives \citep{li-etal-2019-imitation,shi-etal-2019-visually}.
Some of them are tuned with labeled examples in the WSJ development set \citep{shen2018neural,shen2019ordered,htut-etal-2018-grammar,drozdov-etal-2019-unsupervised,kim-etal-2019-compound,wang-etal-2019-tree} or other labeled examples \citep{jin-etal-2018-unsupervised,jin-etal-2019-unsupervised}.

\paragraph{Data augmentation.}
Data augmentation is a strategy for automatically increasing the amount and variety of data for training models, without actually collecting any new data.
Such methods have been found helpful on many NLP tasks, including text classification \citep{kobayashi-2018-contextual,samanta-etal-2019-improved}, relation classification \citep{xu-etal-2016-improved}, and part-of-speech tagging \citep{sahin-steedman-2018-data}. 
Part of our approach also falls into the category of data augmentation, applied specifically to constituency parsing from very few examples. 

\paragraph{Few-shot parsing.}
\citet{sagae-etal-2008-evaluating} show that a supervised dependency parsing model trained on 100 examples can work surprisingly well. 
Recent work has demonstrated the potential of few-shot dependency parsing on multiple languages \interalia{aufrant-etal-2018-quantifying,meechan-maddon-nivre-2019-parse,vania2019systematic}. 
Our approach (\cref{sec:fsp}) can be viewed as few-shot constituency parsing.

\section{Few-Shot Constituency Parsing}
\label{sec:fsp}
We apply Benepar \citep[\cref{sec:benepar};][]{kitaev-klein-2018-constituency} as the base model for few-shot parsing. 
We present a simple data augmentation method (\cref{sec:sub}) and an iterative self-training strategy (\cref{sec:self-training}) to further improve the performance. 
We suggest that such an approach should serve as a strong baseline for unsupervised parsing with supervised tuning.
\subsection{Parsing Model}
\label{sec:benepar}
The Benepar parsing model consists of (i) word embeddings, (ii) transformer--based \citep{vaswani2017attention} word-span embeddings, and (iii) a multi-layer perceptron to compute a score for each labeled span.\footnote{In this work, there are only two labels: (i) NT denotes a constituent and (ii) $\emptyset$ denotes non-constituent. The label $\emptyset$ enables the parser to output non-binary trees; details can be found in \citet{kitaev-klein-2018-constituency}. Almost all existing unsupervised parsing models do not use the nonterminal categories in the development set, so we propose to train such unlabeled constituency parsing models as their baselines. }
The score of an arbitrary tree is defined as the sum of all of its internal span scores.
Given a sentence and its ground-truth parse tree $T^*$, the model is trained to satisfy $\textit{score}(T^*) \geq \textit{score}(T) + \Delta(T^*, T)$ for any tree $T~(T \neq T^*)$, where $\Delta$ denotes the Hamming loss on labeled spans.
The label-aware CKY algorithm is used to obtain the tree with the highest score. 
More details can be found in \citet{kitaev-klein-2018-constituency}. 

\subsection{Data Augmentation}
\label{sec:sub}
\begin{figure}[t]
\flushleft \small \textit{Original sentences:} \\
\begin{subfigure}{0.25\textwidth}
~~~~~~~~~~~~~~~~~~~~~\resizebox{\textwidth}{0.07\textwidth}{
\begin{forest}
 [ \large NT
    [ \large NT
        [ \textit{\large a}] [ \textit{\large cat}]
    ]
    [ \large NT,tikz={\node [draw,dkgreen,inner sep=0,fit to=tree]{};}
        [\textit{\large is}] [\textit{\large drinking}] [\textit{\large milk}]
    ]
 ]
\end{forest}
}
\end{subfigure}
\begin{subfigure}{0.39\textwidth}
\resizebox{\textwidth}{0.043\textwidth}{
\begin{forest}
[\large NT
    [\large NT,tikz={\node [draw,red,dotted,inner sep=0,fit to=tree]{};}
        [\textit{\large several}] [\textit{\large kittens}]
    ]
    [\large NT
        [\textit{\large were}] [\textit{\large born}]
        [\large NT,tikz={\node [draw,blue,dashed,inner sep=0,fit to=tree]{};}
            [\textit{\large in}]
            [\large NT
                [\textit{\large the}] [\textit{\large shelter}]
            ]
        ]
    ]
]
\end{forest}
}
\end{subfigure}

\flushleft \small \textit{Generated sentences:} \\
\centering
\begin{subfigure}{0.23\textwidth}
\resizebox{\textwidth}{0.078\textwidth}{
\begin{forest}
 [ \large NT
    [ \large NT
        [ \textit{\large a}] [ \textit{\large cat}]
    ]
    [ \large NT,tikz={\node [draw,red,dotted,inner sep=0,fit to=tree]{};}
        [ \textit{\large several}] [ \textit{\large kittens}]
    ]
 ]
\end{forest}
}
\end{subfigure}
\begin{subfigure}{0.23\textwidth}
\resizebox{\textwidth}{0.075\textwidth}{
\begin{forest}
 [ \large NT
    [ \large NT
        [ \textit{\large a}] [ \textit{\large cat}]
    ]
    [\large NT,tikz={\node [draw,blue,dashed,inner sep=0,fit to=tree]{};}
        [\textit{\large in}]
        [\large NT
            [\textit{\large the}] [\textit{\large shelter}]
        ]
    ]
 ]
\end{forest}
}
\end{subfigure}
\caption{\label{figure:teaser} 
Illustration of the proposed data augmentation approach for improving few-shot parsing: we create new sentences by subtree substitution (e.g., substituting the subtree in the solid box by the ones in the dotted or dashed box), whether the created sentences are grammatical or not. NT denotes nonterminal nodes. 
}
\end{figure}
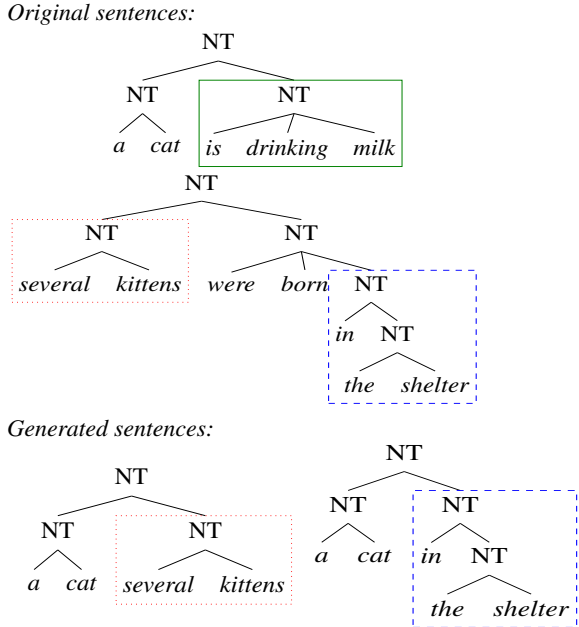
We introduce a data augmentation method, subtree substitution (\sub; Figure~\ref{figure:teaser}), to automatically improve the diversity of data in the few-shot setting. 

We start with a set of sentences with $N$ unlabeled parse trees $\mathcal{S} = \{\langle s_i, \mathcal{T}_i \rangle \}_{i=1}^{N}$; $s_i =\langle w_{i1}, w_{i2}, \ldots, w_{iL_i} \rangle$ denotes a sentence with $L_i$ words, where $w_{ik}$ denotes a word; $\mathcal{T}_i=\{\langle b_{ij}, e_{ij}\rangle \}_{j=1}^{C_i}$ denotes the unlabeled parse tree of $s_i$ with $C_i$ nonterminal nodes; $b_{ij}$ and $e_{ij}$ denotes the beginning and ending index of a constituent.

The augmented dataset $\mathcal{S}'$ is initialized to $\mathcal{S}$.
At each step, we draw a sentence $s_i$ and its parse tree $\mathcal{T}_i$ uniformly from $\mathcal{S}'$, and draw a constituent $\langle b_{ij}, e_{ij}\rangle \in \mathcal{T}_i$ uniformly from $\mathcal{T}_i$. 
After that, we replace $\langle b_{ij}, e_{ij}\rangle$ with a random $\langle b_{kh}, e_{kh}\rangle \in t_k$; that is, we replace a constituent with another one from the training set. 
We let $s_i'$ and $\mathcal{T}_i'$ denote modified sentence and its parse tree, assign $\mathcal{S}' \leftarrow \mathcal{S}' \cup \{(s_i', \mathcal{T}_i')\}$, and repeat the above procedure until $\mathcal{S}'$ reaches the desired size. 

\subsection{Self-Training}
\label{sec:self-training}
\citet{steedman-etal-2003-bootstrapping}, \citet{reichart-rappoport-2007-self} and \citet{mcclosky-etal-2006-effective} have shown that self-training (\st) on unseen sentences can improve a parsing model. 
Inspired by this, we apply an iterative self-training strategy after obtaining each supervised or unsupervised parsing model. 

Concretely, we start with an arbitrary parsing model $\mathcal{M}_0$. 
At the $i^\textrm{th}$ step of self-training, we (i) use the trained model from the previous step (i.e., $\mathcal{M}_{i-1}$) to predict parse trees for sentences in the WSJ training set and those in the WSJ development set, and (ii) train a supervised parsing model $\mathcal{M}_i$ \citep{kitaev-klein-2018-constituency} to fit the prediction of $\mathcal{M}_{i-1}$.
No gold labels are used in self-training.  

\section{Experiments}
\subsection{Dataset and Training Details}
We use the WSJ portion of the Penn Treebank corpus \citep{marcus-etal-1993-building} to train and evaluate the models, replace all number tokens with a special token, and split standard train/dev/test sets following \citet{kim-etal-2019-unsupervised}.\footnote{
For analysis purposes (\cref{sec:st-results} and Figure~\ref{fig:sketch}), we use WSJ Section~24, instead of the standard development set (Section~22) as we train few-shot parsing on part of it. We do not use the standard test split (Section~23) to avoid tuning on the test set, hence our analysis numbers are not directly comparable with those reported in original papers.}
For each criterion, we tune the hyperparameters of each model with respect to its performance on the development set.
To solve the problem of vocabulary sparsity in the few-shot parsing setting (\cref{sec:fsp}), we initialize the word embeddings of Benepar \citep{kitaev-klein-2018-constituency} with the word embeddings from an LSTM--based \citep{hochreiter1997long} language model trained on the WSJ training set. 
During training, models are able to access all sentences (without parse trees) in the WSJ training set; for few-shot parsing or unsupervised parsing with supervised tuning, some unlabeled parse trees in the WSJ development set are available as well. We augment the training set to 10,000 examples for few-shot parsing with \sub, and apply 5-step self-training when applicable. 

We evaluate the unlabeled $F_1$ score of all models using {\tt evalb},\footnote{\url{https://nlp.cs.nyu.edu/evalb/}} discarding punctuation.
More details can be found in the supplementary material. 

\subsection{Models and Tuning Criteria}
\label{sec:criteria}
We investigate four recently proposed models: PRPN \citep{shen2018neural}, ON-LSTM \citep{shen2019ordered}, DIORA \citep{drozdov-etal-2019-unsupervised}, and Compound PCFG \citep{kim-etal-2019-compound}.

PRPN and ON-LSTM are left-to-right neural language models, where syntactic distance \citep{shen-etal-2018-straight} between consecutive words is computed from the model output and used to infer the constituency parse tree. DIORA learns text-span representations and span-level scores by optimizing a masked language modeling objective. The compound PCFG uses a neural parameterization of a PCFG, as well as a per-sentence latent vector which introduces context sensitivity. Both DIORA and the Compound PCFG use the CKY algorithm to infer the parse tree of a given sentence. 

As fully unsupervised tuning criteria, we use perplexity on the development set for PRPN and ON-LSTM, and the upper bound of perplexity for the Compound PCFG, following \citet{shen2018neural,shen2019ordered} and \citet{kim-etal-2019-compound} respectively. For DIORA, we use its reconstruction loss on the development set.\footnote{\citet{drozdov-etal-2019-unsupervised} did not evaluate any unsupervised tuning criteria for DIORA. We choose reconstruction loss because it is what DIORA minimizes during training.} 

\subsection{Comparison between Unsupervised Parsing and Few-Shot Parsing}
\begin{figure}[t]
    \centering
    \includegraphics[width=0.5\textwidth, height=1.55in]{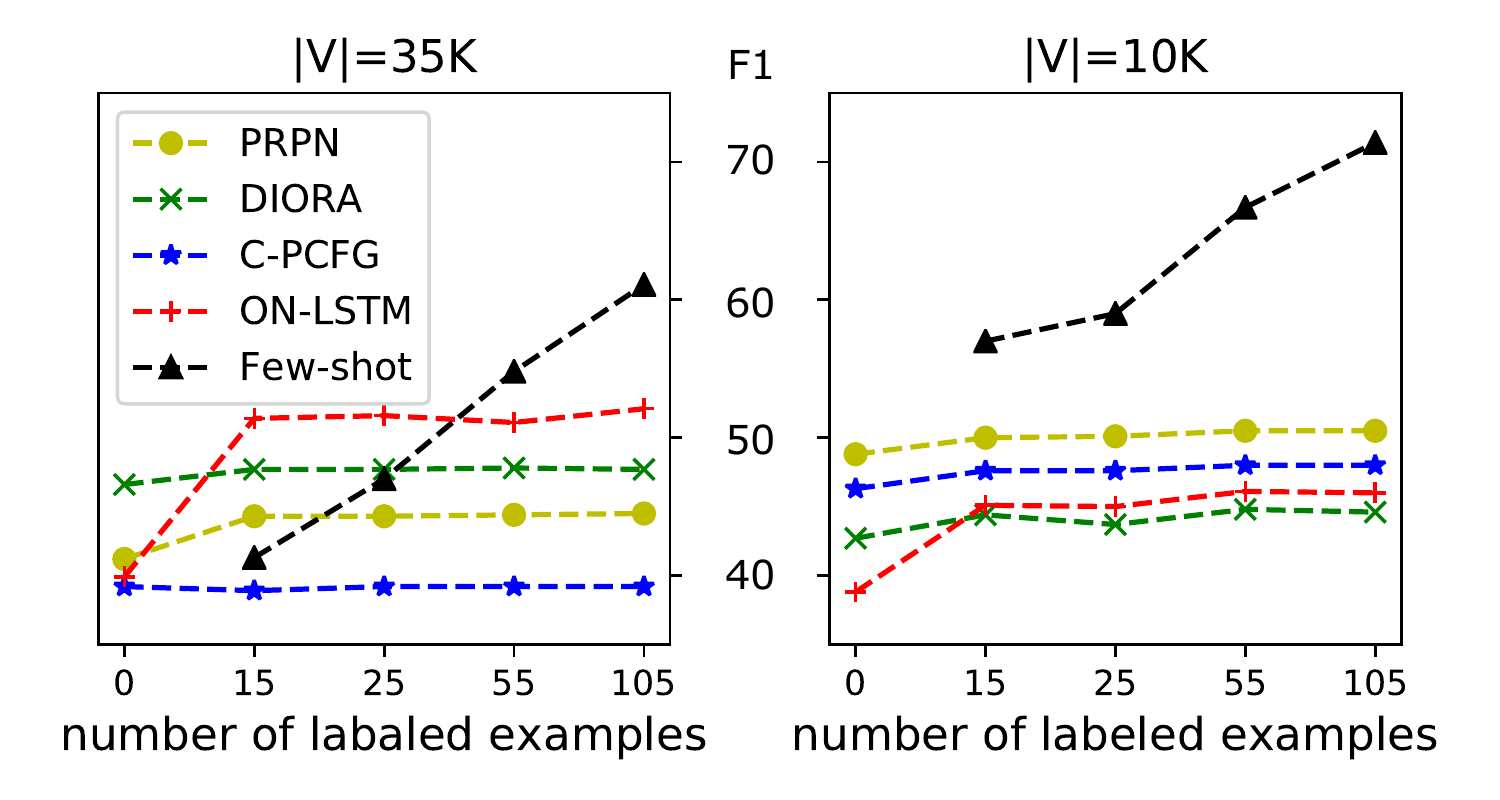} 
    \caption{
    Performance of models with vocabulary size 35K (left) and 10K (right) on WSJ Section 24.
    C-PCFG denotes the Compound PCFG.  The $F_1$ scores are averaged over 5 runs with the same hyperparameters, different random seeds, and different sets of labeled examples when applicable.
    }
    \label{fig:sketch}
\end{figure}

We compare unsupervised parsing against few-shot parsing (Table~\ref{tab:main-results} and Figure~\ref{fig:sketch}): when there are 55 or more labeled examples available, few-shot parsing (\cref{sec:fsp}) consistently outperforms all unsupervised parsing models; with \sub\ and self-training or a smaller vocabulary size ($|V|=10K$), few-shot parsing begins to dominate even when only 15 labeled examples are available. 

On the other hand, we find that a few labeled examples are consistently helpful for most models to achieve better results than fully unsupervised parsing. 
In addition, models tuned on a very small number (e.g., 15) of labeled examples can achieve similar performance to those tuned on 1,700 labeled examples; that is, we need far fewer labeled examples than existing unsupervised parsing approaches have used to obtain very similar results. 

To test if \sub\ can also help improve unsupervised parsing models, we generate 10K sentences from the 1,700 sentences in the WSJ development set with \sub\ (Figure~\ref{figure:teaser}), and add them to the 40K-sentence WSJ training set. We compare unsupervised parsing models trained on the original WSJ training set and the augmented one (Table~\ref{tab:sub-up}). We find that \sub\ can sometimes help, but not by a large margin, and all numbers in Table~\ref{tab:sub-up} are far below the performance of few-shot parsing with the same data availability (82.6; Table~\ref{tab:main-results}). Few-shot parsing with data augmentation is a strong baseline for unsupervised parsing with data augmentation. 

\begin{table}[t]
    \centering 
    \small 
    \begin{tabular}{lcccc}
    \toprule
        \bf Model $ \backslash |\mathcal{D}_\textit{label}|$ & 0 & 15 & 55 & 1,700\\
    \midrule
        PRPN & 42.4 & 44.6 & 44.7 & 44.9 \\
        DIORA & \bf 46.6 & 47.7 & 47.6 & 48.0 \\
        Compound PCFG & 39.2 & 39.2 & 39.2 & 39.2 \\
        ON-LSTM & 39.0 & \bf 51.5 & \bf 51.1& \bf 52.0 \\
    \midrule
        Few-Shot & \na & 42.1 & 55.5 & 81.2 \\
        Few-Shot + \sub &\na & 52.5 & 58.5 & 82.6 \\
        Few-Shot + \sub\ + \st & \na & \bf 53.4 & \bf 61.2 & \bf 85.1\\
    \bottomrule
    \end{tabular}
    \caption{
    Unlabeled $F_1$ scores on the standard WSJ test set (Section~23).
    We keep all tokens, resulting in a vocabulary size $|V|=35K$. $|\mathcal{D}_\textit{label}|=0$ means using fully unsupervised criteria (\cref{sec:criteria}); otherwise we use the first $|\mathcal{D}_\textit{label}|$ labeled examples in WSJ Section 22. 
    For few-shot parsing (\cref{sec:fsp}), we divide the available labeled examples into 10/5, 50/5, and 1,600/100 respectively for training and development. 
    We use boldface for the best unsupervised parsing result and the best few-shot parsing result in each column.  
    }
    \label{tab:main-results}
\end{table}
\begin{table}[]
    \centering \small
    \begin{tabular}{lcc}
        \toprule
         \bf Model & $\text{WSJ}_\textit{train}$ & + $\text{WSJ}_\textit{dev}$ \sub \\
         \midrule
         PRPN & 44.9 & \bf 46.1 \\
         DIORA & 48.0 & \bf 48.2 \\
         Compound PCFG & 39.2 & \bf 42.2 \\
         ON-LSTM & \bf 52.0 & 48.2 \\
         \bottomrule
    \end{tabular}
    \caption{Unlabeled $F_1$ scores on the standard WSJ test set. $\text{WSJ}_\textit{train}$ denotes models trained with the 40K sentences in the WSJ training set, and + $\text{WSJ}_\textit{dev}$ \sub\  denotes models trained with the union of WSJ training sentences and 10K sentences augmented from 1,700 WSJ development sentences. The best number in each row is bolded. }
    \label{tab:sub-up}
\end{table}

\subsection{The Importance of Vocabulary Size}
\label{sec:vocab-size}
We notice that the result of the Compound PCFG in Table~\ref{tab:main-results} is much worse than that reported by \citet{kim-etal-2019-compound}.\footnote{Our DIORA result 
also differs from that reported by \citet{drozdov-etal-2019-unsupervised}; 
however, our number is not directly comparable to theirs due to different data settings---they use a different training set and apply ELMo \citep{peters-etal-2018-deep} for model initialization. 
}
The only major difference between their approach and ours is the vocabulary size: instead of keeping all words, they keep the most frequent 10K words in the WSJ corpus and replace others with a special token. 
To analyze the importance of this choice, we compare the performance of the models with vocabulary size 35K vs.~10K (Figure~\ref{fig:sketch}), tuning models separately in the two settings.
We find that the vocabulary size, which has not been widely considered a hyperparameter and varies across prior work, greatly affects the performance of all models tested.
One possible reason is that a large portion (79.9\%) of the low-frequency (i.e., outside the 10K vocabulary) word tokens are nouns or adjectives -- some models (e.g., PRPN and Compound PCFG) may benefit from collapsing these tokens to a single form, as it may be a beneficial kind of word clustering.
This suggests that we should consider tuning the vocabulary size as a hyperparameter, or fix the vocabulary size for fair comparison in future work.  

\subsection{Self-Training Improves all Models}
\label{sec:st-results}
Inspired by the fact that self-training boosts the performance of few-shot parsing (Table~\ref{tab:main-results}), we apply iterative self-training to the unsupervised parsing models as well, and find that it improves all models (Table~\ref{tab:self}).\footnote{A similar idea and similar results have been presented by \citet{kim-etal-2019-compound}, where they train an RNNG \cite{dyer-etal-2016-recurrent} to fit the prediction of unsupervised parsing models. 
} 
It is worth noting that 5-step self-training is better than 1-step self-training for all base models we experimented with. 
Our results suggest that iterative (e.g., 5-step) self-training may be considered as a standard post-hoc processing step for unsupervised parsing. 
\begin{table}[]
    \centering
    \small
    \begin{tabular}{lccc}
        \toprule
         \bf Model & \multicolumn{3}{c}{\bf \#\st-steps} \\
         & 0 & 1 & 5 \\
        \midrule
         PRPN & 44.7 & 44.7 & 45.1\\
         DIORA & 46.7 & 48.7 & 49.1 \\
         Compound PCFG & 41.1 & 41.8 & 42.2 \\
         ON-LSTM & 50.2 & 51.3 & 52.1 \\
         Few-Shot & 44.3 & 44.5 & 45.0 \\
         Few-Shot + \sub & 53.3 & 55.5 & 56.6 \\
        \bottomrule
    \end{tabular}
    \caption{$F_1$ score on WSJ Section 24 of different models, where the base models are those used to report results in Table~\ref{tab:main-results} with $|\mathcal{D}_\textit{label}|=15$. }
    \label{tab:self}
\end{table}
\section{Discussion}
While many state-of-the-art unsupervised parsing models are tuned on all labeled examples in a development set \interalia{drozdov-etal-2019-unsupervised,kim-etal-2019-unsupervised,wang-etal-2019-tree}, we have demonstrated that, given the same data, few-shot parsing with simple data augmentation and self-training can consistently outperform all of these models by a large margin.  
We suggest that one possibility for future work is to focus on fully unsupervised criteria, such as language model perplexity \citep{shen2018neural,shen2019ordered,kim-etal-2019-unsupervised,peng-etal-2019-palm,li-etal-2020-empirical} and model stability across different random seeds \citep{shi-etal-2019-visually}, for model selection, as discussed in unsupervised learning work \interalia{smith-eisner-2005-contrastive,smith-eisner-2006-annealing,spitkovsky-etal-2010-baby,spitkovsky-etal-2010-viterbi}. 
An alternative is to use as few labeled examples in the development set as possible, and compare to few-shot parsing trained on the used examples as a strong baseline.
In addition, we find that self-training is a useful post-processing step for unsupervised parsing. 

Our work does not necessarily imply that unsupervised parsers produce poor parses; they may be producing good parses that clash with the conventions of treebanks \citep{klein2005unsupervised}. 
If this is the case, then extrinsic evaluation of parsers in downstream tasks \citep{shi-etal-2018-tree}, e.g., machine translation \citep{denero-uszkoreit-2011-inducing,neubig-etal-2012-inducing,gimpel-smith-2014-phrase}, may better show the potential of unsupervised methods.

\section*{Acknowledgments}
We thank Allyson Ettinger, Yoon Kim, Jiayuan Mao, Shane Settle, and Shubham Toshniwal for helpful discussions, as well as all the knowledgeable anonymous reviewers for their valuable and insightful feedback. 

\bibliographystyle{acl_natbib}
\bibliography{anthology,emnlp2020}

\end{document}